\documentclass[10pt,twocolumn,letterpaper]{article}
\usepackage[pagenumbers]{cvpr}
\usepackage{amsmath}
\usepackage{amssymb}
\usepackage{booktabs}
\usepackage{xcolor}
\usepackage{colortbl}
\usepackage{xspace}
\usepackage{tabularx}
\usepackage{multirow}
\usepackage{array}
\usepackage{makecell}
\usepackage{mathtools}
\usepackage{microtype}

\def\redc{\bfseries\cellcolor[HTML]{FCE4E4}}
\def\orangec{\cellcolor[HTML]{FFF0E0}}
\def\yellowc{\cellcolor[HTML]{FFFADE}}

\providecommand{\eg}{e.g.\@\xspace}
\providecommand{\Eg}{E.g.\@\xspace}
\providecommand{\ie}{i.e.\@\xspace}

\providecommand{\etc}{etc.\@\xspace}

\newcommand{\methodname}{GalaxyGS}

\usepackage{stfloats}
\definecolor{paperblue}{rgb}{0.21,0.49,0.74}
\usepackage[breaklinks,colorlinks,allcolors=paperblue,pdfusetitle]{hyperref}
\title{Multi-Band Galaxy Novel-View Synthesis with Radiative Gaussian Fields}
\author{%
  Yuzhe Wu\textsuperscript{1}, Jialin Zhu\textsuperscript{2},
  Do\u{g}a Y{\i}lmaz\textsuperscript{1},
  Daisuke Kawata\textsuperscript{1}, He Wang\textsuperscript{1,3}%
  \thanks{Corresponding author, \href{mailto:he_wang@ucl.ac.uk}{\nolinkurl{he_wang@ucl.ac.uk}}}\\
  \textsuperscript{1}University College London, UK\quad
  \textsuperscript{2}Baidu\\
  \textsuperscript{3}AI Centre, University College London, UK%
}
\hypersetup{pdfauthor={Yuzhe Wu, Jialin Zhu, Do\u{g}a Y{\i}lmaz, Daisuke Kawata, He Wang}}

\begin{document}
\maketitle
\begin{abstract}
We propose a novel view synthesis tool for galaxies based on Gaussian Splatting (GS). Unlike RGB data, where GS has achieved success, galaxy observations consist of multi-band signals from a massive number of bodies, contain complex patterns of cross-band interactions, and capture light propagation in highly heterogeneous media. As a result, the underlying physical process poses great challenges in model representation and optimization, which have not been investigated in existing research. To this end, we propose a physically inspired Gaussian Splatting method, GalaxyGS, which is equipped with physically meaningful components that capture both the energy emission mechanism of different astronomical bodies and the light propagation between them. In the absence of directly comparable baseline methods, we adopt a wide range of possible alternative approaches. Comparisons conclusively demonstrate that GalaxyGS is better in accuracy and physical fidelity across datasets and metrics. 
\end{abstract}

\section{Introduction}
\label{sec:intro}

3D Gaussian Splatting (3DGS)~\cite{kerbl2023gaussian} has become a versatile foundational component for 3D reconstruction, simultaneous localization and mapping~\cite{keetha2024splatam}, autonomous driving~\cite{yan2024street}, and 3D representations in generative models~\cite{yi2024gaussiandreamer}. So far, its applications have often focused on RGB and RGB-D data, sometimes with additional appearance or geometric information such as texture, lighting, and surface normals~\cite{liang2024gsir}. In other words, the synthesized novel view is often restricted to the visible spectrum of human eyes. Recently, 3DGS has been extended to data that go beyond the visible spectrum, such as thermal imaging~\cite{lu2025thermalgaussian} and hyperspectral reconstruction~\cite{thirgood2025hypergs}. Our research also extends 3DGS into a new domain, multi-band galaxy novel view synthesis, where the signals cover a wide range of wavelengths with light propagation through complex and heterogeneous media. Given sparse multi-band images of a galaxy, we reconstruct galaxies by predicting multi-band images at unseen viewpoints, extending 3DGS into a scientific research tool for astronomy.

Galaxy reconstruction poses three coupled representational and optimization challenges, which demand a fundamental modeling innovation rather than a naive adaptation of existing 3DGS research. The first challenge is the heterogeneous media. A galaxy observation contains mixed multi-band signals from stars and dust of various sizes and properties. This gives compound signals which, if not untangled well, severely compromise novel view synthesis. These components have distinct emission mechanisms, manifested as different energy distributions across wavelengths, \eg ultraviolet-to-near-infrared (UV-NIR), mid-infrared (MIR), and far-infrared to submillimeter (FIR/submm) ranges~\cite{conroy2013panchromatic}. The energy of stellar emission is concentrated in the UV-NIR range, while dust emission extends across the MIR and FIR/submm ranges~\cite{draine2007dust}. This is significantly different from existing methods where stellar and dust emission mechanisms are rarely explicitly considered, even in those designed for spectral signals~\cite{thirgood2025hypergs,meyer2026mssplatting} and astronomy~\cite{levis2022bhnerf,jarolim2024sunerf,nolan2026astrosplat}.

The immediately following challenge is the complex light propagation in heterogeneous media. Stellar emission is strong but attenuated by dust through absorption and scattering, and absorbed energy heats the dust, which further re-emits energy mainly in the infrared range, producing a thermal dust continuum~\cite{steinacker2013dust}. At the same time, there are small dust grains that also absorb and re-emit energy, contributing strongly to emission in the MIR range~\cite{draine2007dust}. Existing 3DGS research, including spectral models, often simplifies the propagation process without modeling the behaviors of different media during propagation~\cite{meyer2026mssplatting}. In our case, not only do we need to model emission and its propagation, but we also aim to make them physically meaningful, \eg attributing the correct amount of energy to different components when learning from mixed signals.

Last but not least, heterogeneous media and complex light propagation jointly impose challenges on optimization, which needs to accurately synthesize views and simultaneously untangle the signals and attribute them to corresponding components. Such optimization cannot be achieved by simple stochastic gradient~\cite{kerbl2023gaussian} or sampling methods~\cite{kheradmand2024mcmc} alone, without a representation that distinguishes emission from attenuation, as shown later.

To jointly address these challenges, we propose a new 3DGS method named Galactic Gaussian Splatting (GalaxyGS). GalaxyGS is physically inspired and consists of two key elements: structured primitives and staged optimization. We first create three Gaussian fields in 3D space: a stellar emission field, a dust attenuation field, and a dust emission field. We design three types of primitives for these fields. Next, the first two fields jointly model stellar emission and its attenuation, while the last field captures the dust re-emission. These physically inspired fields explicitly model light propagation and assign distinct emission and attenuation contributions to the observed multi-band signals. Having addressed the representational challenges, we find that jointly training all components easily suffers from local minima. This involves not only a highly nonconvex optimization problem but also solution ambiguity, \ie multiple physical models can give the same observation~\cite{denicola2020deprojection,kouros2024ambiguity}. Therefore, we propose a new physically inspired strategy for staged optimization of the fields.

We evaluate \methodname{} on two datasets: Auriga~\cite{kapoor2021auriga} and TNG50~\cite{costantin2023ceers}. Since there are no GS methods for galaxy reconstruction, we adapt six closely related methods, which cover a wide range of representations and optimization strategies. Evaluation conclusively demonstrates the performance of \methodname{} across galaxies of different types, formation mechanisms, \etc. \methodname{} universally outperforms the best baseline method by $31.4\%$ in reconstruction accuracy and $80.2\%$ in physical fidelity. Our contributions include:

\begin{itemize}
    \item the first 3DGS method for multi-band galaxy novel view synthesis, to the best of our knowledge.
    \item a physically inspired 3DGS representation that explicitly models energy emission mechanisms.
    \item a physically inspired light propagation model for 3DGS.
    \item a staged optimization strategy to enable 3DGS to disentangle mixed signals.
\end{itemize}

\section{Related Work}
\label{sec:related}

\noindent{\bf Multi-band representations.} Multi-band view synthesis extends appearance modeling beyond RGB through wavelength-dependent radiance or shared spectral features. HS-NeRF~\cite{chen2024hsnerf} conditions radiance on wavelength, while ThermalGaussian~\cite{lu2025thermalgaussian} couples RGB and thermal Gaussians through multimodal regularization. Using shared spectral features, HyperGS~\cite{thirgood2025hypergs} renders in a pretrained autoencoder's latent space and decodes the result into spectra. MS-Splatting~\cite{meyer2026mssplatting} instead decodes shared per-Gaussian features into band radiance before rendering. For direct spectral modeling, DD-HGS~\cite{narayanan2025ddhgs} conditions spherical harmonic coefficients on wavelength and denoises rendered spectra with diffusion. Material-based approaches give a physical interpretation. UnMix-NeRF~\cite{perez2025unmixnerf} models diffuse spectra as nonnegative mixtures of learned material spectra, and SpectralGaussians~\cite{sinha2024spectralgaussians} separates spectral reflectance and illumination for relighting. These methods advance spectral modeling from appearance prediction toward physical interpretation, but do not explicitly distinguish objects with different emission mechanisms.

\noindent{\bf Neural rendering with complex media.} The observed image depends on source spectra and light propagation through the scene. FitSKIRT~\cite{degeyter2013fitskirt} estimates parametric stellar and dust distributions by comparing SKIRT radiative-transfer simulations with galaxy images. Astronomical neural renderers incorporate domain-specific propagation. BH-NeRF~\cite{levis2022bhnerf} recovers emission near black holes using gravitationally bent rays, while SuNeRF~\cite{jarolim2024sunerf} models solar-coronal emission and absorption. For small-body surfaces, AstroSplat~\cite{nolan2026astrosplat} instead ties Gaussian intensity to surface orientation and sunlight through planetary reflectance models. SeaThru-NeRF~\cite{levy2023seathrunerf} and SeaSplat~\cite{yang2025seasplat} similarly separate scene appearance from wavelength-dependent attenuation and backscatter in water. These neural renderers model specialized propagation effects, but not the distinct emission mechanisms and attenuation that coexist in galaxy images.

\noindent{\bf Optimization.} Other than gradient-based optimization with adaptive cloning and splitting~\cite{kerbl2023gaussian} or sampling~\cite{kheradmand2024mcmc}, existing research uses bespoke optimization for different scenarios. Under sparse views, LGDWT-GS~\cite{salehi2026lgdwt} adds global and patch-wise image wavelet losses to regularize scene fitting. 3DSatellites~\cite{chang2025satellites} couples Gaussian reconstruction with camera-pose estimation. Estimating physical properties introduces a different difficulty, because material and illumination changes can compensate in the rendered image~\cite{kouros2024ambiguity}. GS-IR~\cite{liang2024gsir} organizes this estimation in stages, reconstructing geometry and caching occlusion before fitting material properties and illumination. In our research, separating mixed radiative signals requires controlling compensation between emission and attenuation.

\section{Preliminaries}
\label{sec:preliminaries}

\begin{figure*}[tb]
  \centering
  \includegraphics[width=0.94\textwidth]{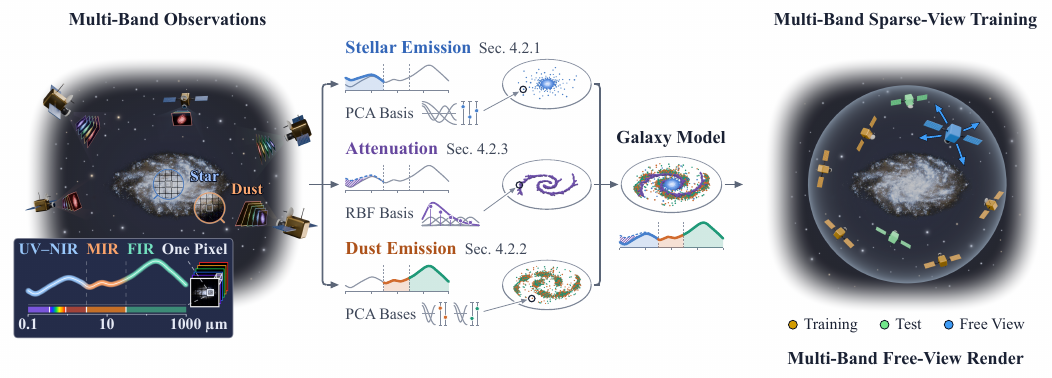}
 \caption{Overview of \methodname{}. Multi-band observations provide a sampled spectrum at each pixel across the UV--NIR, MIR, and FIR/submm bands (left). We represent stellar emission, dust attenuation, and dust emission with separate Gaussian fields, using PCA bases for emission spectra and an RBF basis for wavelength-dependent attenuation (middle). The combined galaxy model is fitted to sparse multi-band views and renders images in the same bands from novel viewpoints (right).}
\label{fig:method-overview}
\end{figure*}

A typical galaxy observation (\cref{fig:method-overview} left)
is a multi-band image $\mathbf I\in\mathbb R^{h\times w\times C}$,
where $h$, $w$, and $C$ denote the image height, width, and number of bands (channels), respectively. 
This is a discretized representation in both space and wavelength. The raw data at a pixel form a continuous spectrum of wavelengths. It is then discretized into $C$ bands. 
At each pixel, band $c\in\{1,\ldots,C\}$ contains light signals over a range of wavelengths and is labeled by a representative
wavelength $\lambda_c$. 
Furthermore, we divide the bands into three band groups: UV-NIR below $3\,\mu\mathrm m$, MIR from $3$ to $30\,\mu\mathrm m$, and FIR/submm beyond $30\,\mu\mathrm m$~\cite{hickox2018obscured}. 

Galaxy observations are more complex than the RGB data in standard GS, and novel view synthesis for galaxies requires physical fidelity on top of visual realism. As a result, cross-band interactions need to be explicitly considered and physically meaningful measures of light distribution are required, none of which is needed in standard GS.

Next, different astronomical bodies show distinctive behaviors in light emission and attenuation. Their emissions can span the whole spectrum but concentrate on different band groups, shown in~\cref{fig:method-overview} left. First, stellar emission dominates the UV-NIR group. Its spectral shape varies with stellar age and metallicity~\cite{bruzual2003stellar}. Hotter stars, for example, emit a greater fraction of their light at shorter wavelengths. Dust absorbs and scatters stellar light, causing dimming. This intensity of the dimming depends on the optical depth~\cite{steinacker2013dust} and the wavelength, also described by a function called an extinction curve~\cite{fitzpatrick1999extinction}. Furthermore, dust is heated by absorbed light and then emits in the IR groups, where small grains contribute strongly in the MIR group, while larger grains dominate the FIR/submm group~\cite{draine2007dust}. 

\section{Methodology}
\label{sec:method}

We propose \methodname{}, a physically inspired model of light propagation in galaxies (\cref{fig:method-overview}). It represents stellar emission, dust attenuation, and dust emission with separate Gaussian fields, fitted through a specially designed optimization. One key design is to constrain the learning of different Gaussian fields within physically meaningful spaces spanned by appropriate basis functions.

\subsection{Gaussian Fields}
\label{sec:gaussian-fields}
\label{sec:image-formation}
\label{sec:joint-radiative-compositing}
\label{sec:training-composition}
We model a galaxy as a distribution of stars and dust, and explicitly represent emission and attenuation with three Gaussian fields: 
\begin{equation}
\mathcal G
=
\mathcal G_{\mathrm{star}}
\cup\mathcal G_{\mathrm{dust}}
\cup\mathcal G_{\mathrm{att}},
\label{eq:gaussian-fields}
\end{equation}
where $\mathcal G_{\mathrm{star}}$, $\mathcal G_{\mathrm{dust}}$, and $\mathcal G_{\mathrm{att}}$ represent stellar emission, dust emission, and dust attenuation, respectively. Stars and dust have different spatial distributions and radiative roles~\cite{galliano2018dust}. Stars are direct sources of strong emission. Dust emission only happens when it absorbs stellar emission, \ie an indirect source of emission, and this light absorption causes attenuation. Dust attenuation is complex and depends on the spatial distribution of the dust with respect to stars and cameras. We use Gaussians parameterized by a mean $\boldsymbol\mu$, covariance $\boldsymbol\Sigma$, view-independent emission vector $\mathbf e\in\mathbb R^C$ for its $C$ bands, and opacity $\alpha$ for all three fields, and they differ only in how they model emission $\mathbf e$ and opacity $\alpha$.

We use $\mathcal G_{\mathrm{star}}\cup\mathcal G_{\mathrm{att}}$ for the UV-NIR group and $\mathcal G_{\mathrm{dust}}$ for the MIR and FIR/submm groups. Rendering follows 3DGS compositing~\cite{kerbl2023gaussian}, applied separately to each band:
\begin{equation}
\begin{aligned}
I_c(\mathbf p)
=\sum_i T_{i,c}(\mathbf p)\alpha_{i,c}(\mathbf p)e_{i,c},\\
T_{i,c}(\mathbf p)
=\prod_{j<i}\left(1-\alpha_{j,c}(\mathbf p)\right),
\end{aligned}
\label{eq:attenuation-compositing}
\end{equation}
where $i$ and $j$ index Gaussians sorted from front to back,
and $c$ indexes the band. $I_c(\mathbf p)$ is the predicted light intensity at pixel $\mathbf p$ in band $c$, expressed in the same physical units as the observations. $\alpha_{i,c}(\mathbf p)$ is pixel opacity, and $T_{i,c}(\mathbf p)$ is the transmittance. The attenuation field adds no light and only reduces light from emitting Gaussians. Finally, the whole prediction is a concatenation of the predicted UV-NIR, MIR, and FIR/submm images for view $v$:
\begin{equation}
{\mathbf I}_v
=
\left[
{\mathbf I}^{\text{UV-NIR}}_v,
{\mathbf I}^{\text{MIR}}_v,
{\mathbf I}^{\text{FIR/submm}}_v
\right].
\label{eq:multiband-assembly}
\end{equation}

\subsection{Physically Inspired Gaussians}
\label{sec:emission-representation}
\label{sec:spectral-emission}
\label{sec:emission-factorisation}

\subsubsection{Stellar Gaussians}
Stellar emission dominates the UV-NIR range, but naive learning is unlikely to give good results, because fitting 3D Gaussians from 2D images is ambiguous~\cite[Sec.~5.1]{kerbl2023gaussian}, \ie different sets of Gaussians can produce the same view. This might not be an issue if only visual realism is the goal, while in GalaxyGS, we aim to enforce physical realism. 

A key observation is that stellar emission is featured by its energy distribution across bands. The cross-band covariance is not random, but has patterns that strongly depend on the age and metallicity of the stellar populations~\cite{bruzual2003stellar}. This means that the covariance patterns can serve as a physically meaningful space to represent stellar emission. Therefore, we use it to establish a basis constraining the learning. 

Specifically, we collect $n$ stellar emission spectra $\mathbf X\in\mathbb R^{n\times C}$ with a mean $\mathbf{\bar{X}}$ from galaxies varying in age and metallicity~\cite{bruzual2003stellar}. We then apply Principal Component Analysis (PCA) and retain the top $K$ eigenvectors as the columns of $\mathbf B\in\mathbb R^{C\times K}$.
For a band group with $C$ bands, the emission vector
$\mathbf e\in\mathbb R^C$ is parameterized as
\begin{align}
\mathbf e&=\operatorname{softplus}(a)\mathbf s,  \nonumber\\ 
\text{where }\mathbf s&=\operatorname{softmax}\!\left(
\mathbf{\bar{X}}+\mathbf B\mathbf z\right),
\mathbf z=\boldsymbol\beta\odot\tanh(\mathbf u),
\label{eq:emission-decoder}
\end{align}
where $a$ is a learnable scalar and $\mathbf u\in\mathbb R^K$ is a learnable vector. $\boldsymbol\beta\in\mathbb R_{>0}^K$ is predefined \eg 0.1. 

\cref{eq:emission-decoder} constrains the learning in a physically meaningful space. First, the ultimate emission $\mathbf e$ is dictated by $\mathbf s$ (scaled by $a$) which is the key learning target. Then $\mathbf s$ is based on a PCA reconstruction $\mathbf{\bar{X}}+\mathbf B\mathbf z$ where the only learnable parameter is $\mathbf z$. Essentially, we are learning the PCA scores of our target galaxy in the coordinate system spanned by the eigenvectors of $(\mathbf X-\mathbf{\bar{X}})^{\mathsf T}(\mathbf X-\mathbf{\bar{X}})$. Since these eigenvectors capture the variances of stellar emission from galaxies, this PCA space is naturally physically meaningful. Furthermore, to avoid learning extreme values for $\mathbf z$, we constrain it by learning $\mathbf u$ instead of $\mathbf z$ directly. Since $-1 \leq \tanh(\mathbf u)\leq 1$ and we can control $\boldsymbol\beta$, $\mathbf z$ is constrained.

\subsubsection{Dust Gaussians}
Dust absorbs stellar radiation and re-emits across two band groups, MIR and FIR/submm, corresponding to different dust grain properties and heating~\cite{draine2007dust}. MIR features from small grains vary in strength, while the shape of the FIR/submm thermal spectrum varies with dust temperature.

To make the learning physically inspired, we propose a strategy similar to that for stellar emission. We collect multi-band images from outside our data, referred to as reference images. For each band group, we sample $n$ pixels with positive intensities across its $C$ bands, giving $\mathbf X^{\mathrm{ref}}\in\mathbb R^{n\times C}$.
We then apply PCA to each group's $\mathbf X^{\mathrm{ref}}$, giving the PCA bases of MIR and FIR/submm, respectively. Finally, we learn the emission of both groups using~\cref{eq:emission-decoder}. 
Note that the means and covariances of the dust Gaussians are shared between MIR and FIR/submm, but each group has its own PCA scores $\mathbf z$ and amplitude parameter $a$, allowing its spectral shape and emission amplitude to vary independently. The sharing of means and covariances is because the MIR and FIR/submm signals are mainly from the small and large grains in the same dust. In other words, the dust is a continuum with randomly distributed particles (small and large grains). Empirically, independently learning both with separate means and covariances leads to ill-conditioned optimization.

\subsubsection{Attenuation Gaussians}
\label{sec:dust-attenuation}
\label{sec:attenuation-spectrum}

Dust absorption and scattering reduce the stellar light reaching the camera, with attenuation depending on wavelength and the positions of stars and dust along the viewing path~\cite{steinacker2013dust}. To this end, we parameterize each attenuation Gaussian with its mean $\boldsymbol\mu$, covariance $\boldsymbol\Sigma$, and a nonnegative optical depth $\tau_c$ for each UV-NIR band $c$. Optical depth $\tau_c$ in astronomy measures the attenuation strength: a larger value means less light is transmitted. Note that attenuation Gaussians have zero emission $e_c=0$ as they only attenuate light. Following the transmittance in volume rendering~\cite{mildenhall2020nerf}, we convert optical depth $\tau_c$ to central opacity $1-\exp(-\tau_c)$, where $\exp(-\tau_c)$ is the transmitted fraction. Multiplying this opacity by the projected Gaussian as in 3DGS~\cite{kerbl2023gaussian} gives
\begin{equation}
\alpha_c(\mathbf p)
=\left(1-\exp(-\tau_c)\right)G(\mathbf p),
\label{eq:attenuation-pixel-opacity}
\end{equation}
where $G(\mathbf p)$ is the projected attenuation Gaussian at pixel $\mathbf p$, and $\alpha_c(\mathbf p)$ is the pixel opacity used in~\cref{eq:attenuation-compositing}.

In \cref{eq:attenuation-pixel-opacity}, light attenuation is dictated by $\tau_c$, 
which varies smoothly with wavelength~\cite{fitzpatrick1999extinction}. We therefore treat $\boldsymbol\tau$ as a function of bands and use Gaussian radial basis functions (RBFs) in log wavelength to constrain the learning:
\begin{equation}
\boldsymbol\tau=\mathbf B_\tau\boldsymbol\eta,\qquad
\boldsymbol\eta=\beta_\tau\operatorname{sigmoid}(\mathbf v),
\label{eq:attenuation-optical-depth}
\end{equation}
where $\boldsymbol\tau\in\mathbb R^C$ denotes the optical depths for each band. The attenuation coefficient vector $\boldsymbol\eta$ is parameterized by a learnable vector $\mathbf v\in\mathbb R^{K_\tau}$ and a predefined upper bound $\beta_\tau>0$, with sigmoid applied elementwise. $\mathbf B_\tau\in\mathbb R^{C\times K_\tau}$ contains $K_\tau$ basis functions, with $(\mathbf B_\tau)_{c,k}=\phi_k(\lambda_c)$, and the basis functions are
\begin{equation}
\phi_k(\lambda_c)
=\exp\!\left[-\frac{(\ln\lambda_c-\ln\lambda_k)^2}{2\sigma^2}\right],
\label{eq:attenuation-rbf}
\end{equation}
where $\lambda_k$ is the predefined center wavelength of the $k$-th basis function and $\sigma$ is the predefined standard deviation.

\subsection{Staged Optimization}
\label{sec:component-optimization}
With all types of Gaussians specified, we can train the model. However, different types of Gaussians can compensate for one another when fitting the data. \Eg weak emission plus weak attenuation gives a similar result to strong emission plus strong attenuation. We therefore propose a three-stage optimization below.

\subsubsection{Training Schedule}
\label{sec:fitting-schedule}
Stellar and dust emission dominate different band groups. Therefore, we first fit the stellar emission field and dust emission field independently. With the stellar and dust PCA bases fixed,
each field learns its Gaussian means, covariances, opacity, PCA scores, and amplitudes together. During this learning stage, we adapt Gaussian positions and counts through the relocation, growth, and covariance-scaled position perturbations of 3DGS-MCMC~\cite{kheradmand2024mcmc}.

In the second stage, stellar emission can slightly overestimate the emission because dust attenuation was not explicitly modeled earlier.
To mitigate this, we fix the stellar field and initialize the attenuation field in the same way as the stellar Gaussians. 
We keep the attenuation means and covariances fixed and learn
only $\mathbf v$ (\cref{eq:attenuation-optical-depth}) against
the same training images.
This estimates wavelength-dependent dimming against a fixed stellar field, reducing ambiguity between emission and attenuation. 

In the third stage, we jointly refine $\mathbf{z}$, $a$ in \cref{eq:emission-decoder}, the means of the stellar Gaussians, and attenuation $\boldsymbol\eta$, while keeping the dust emission fixed. This is done by minimizing image and spectral losses, explained later.

\subsubsection{Training Objectives}
\label{sec:field-training-objectives}
{\bf First-stage loss.} The loss function is:
\begin{equation}
\begin{aligned}
\mathcal L_{\mathrm{em}}
={}&\lambda_r\mathcal L_{\mathrm{MSE}_w}
  +\lambda_{\mathrm{c}}\mathcal L_{\mathrm{cos}}\\
&+\lambda_p\operatorname{NMSE}(\mathbf I,\mathbf I^{\mathrm{GT}})
  +\lambda_f\mathcal L_{\mathrm{flux}}+\mathcal L_{\mathrm{reg}}.
\end{aligned}
\label{eq:field-objectives}
\end{equation}
where $\mathbf I,\mathbf I^{\mathrm{GT}}\in\mathbb{R}^{h\times w\times C}$ denote the predicted and ground-truth (GT) images with $C$ bands. MSE and NMSE are mean squared error and normalized MSE, respectively. 
$\Phi$ is a normalized inverse hyperbolic sine (asinh) transform, so that both bright and faint structures in images contribute~\cite{jarolim2024sunerf}.
\begin{align}
\mathcal L_{\mathrm{MSE}_w}
=&\frac{1}{hwC}\sum_h\sum_w\sum_{c=1}^{C}
\left[\mathbf W^{\mathrm b}\odot
  \bigl(\Phi(\mathbf I_c)-\Phi(\mathbf I_c^{\mathrm{GT}})\bigr)^2\right],\nonumber\\
&\text{where } \mathbf W^{\mathrm b}
=(1-\alpha)+\alpha\frac{hw}{N_{\mathrm b}}\mathbf M.
\label{eq:weighted-mse}
\end{align}
where $\mathbf I_c,\mathbf I_c^{\mathrm{GT}}\in\mathbb R^{h\times w}$ are the predicted and GT images in band $c$, $\mathbf W^{\mathrm b}\in\mathbb R^{h\times w}$ is the weight, and $\odot$ denotes elementwise multiplication. Because galaxy observations contain large areas of dark pixels, corresponding to near-zero emission, they can easily overshadow the galaxy itself during training. Therefore, we focus on the bright areas in the MSE. Specifically, $\mathbf M\in\{0,1\}^{h\times w}$ is a mask, $1$ for bright regions in GT and $0$ otherwise. Bright regions are defined as pixels with the top $10\%$ of light intensity in $\Phi(\mathbf I_c^{\mathrm{GT}})$. $N_{\mathrm b}$ is the number of pixels in bright regions, and $\alpha$ is predefined. 

$\mathcal L_{\mathrm{reg}}$ penalizes opacity and spatial standard deviation following 3DGS-MCMC~\cite{kheradmand2024mcmc}. $\mathcal L_{\mathrm{cos}}$ is the cosine spectral loss which is effective for hyperspectral data~\cite{thirgood2025hypergs}.
At each pixel $\mathbf p$, we define a feature $\mathbf y_{\mathbf p}\in\mathbb R^C$, where $y_{\mathbf p,c}=\max(\Phi(I_c(\mathbf p)),0)$, for the prediction and GT:
\begin{align}
\mathcal L_{\mathrm{cos}}
=\sum_{h}\sum_{w} w_{\mathbf p}^{\mathrm{cos}}
\left(
1-
\frac{\mathbf y_{\mathbf p}^{\mathsf T}\mathbf y_{\mathbf p}^{\mathrm{GT}}} 
{\|\mathbf y_{\mathbf p}\|_2
 \|\mathbf y_{\mathbf p}^{\mathrm{GT}}\|_2}
\right) \nonumber, \\
\text{where } w_{\mathbf p}^{\mathrm{cos}} = \frac{\|\mathbf y_{\mathbf p}^{\mathrm{GT}}\|_2}{\sum_{h}\sum_{w}\|\mathbf y_{\mathbf p}^{\mathrm{GT}}\|_2}.
\label{eq:training-cosine}
\end{align}
Intuitively, $\mathcal L_{\mathrm{cos}}$ measures the similarity of two per-pixel spectra without comparing intensities.

To also account for intensities, we use a flux loss for each of the UV-NIR, MIR, and FIR/submm band groups, indexed by $g$:
\begin{align}
&\mathcal L_{\mathrm{flux}}
=\operatorname{MSE}(\widehat{\mathbf f},\mathbf f), \nonumber \\
&\widehat f_g=\log_{10}[\operatorname{sum}(\mathbf I_g)], f_g=\log_{10}[\operatorname{sum}(\mathbf I_g^{\mathrm{GT}})],
\label{eq:flux-loss}
\end{align}
where $\widehat{\mathbf f}$ and $\mathbf f$ concatenate the predicted
and GT fluxes $\widehat f_g$ and $f_g$, respectively,
$\operatorname{sum}$ adds all pixel intensities in $\mathbf I_g$,
and $\log_{10}$ compresses the dynamic range, reducing the dominance
of large values in the loss.

\noindent {\bf Second-stage loss.} With the learned stellar field fixed, we compare its attenuated UV-NIR
prediction $\mathbf I_s$ with the corresponding GT image $\mathbf I_s^{\mathrm{GT}}$:
\begin{equation}
\mathcal L_{\mathrm{att}}
=\lambda_a\operatorname{MSE}\!\left(
  \Phi(\mathbf I_s),\Phi(\mathbf I_s^{\mathrm{GT}})\right).
\label{eq:attenuation-mae}
\end{equation}

\noindent {\bf Third-stage loss.} For joint refinement, we could use \cref{eq:field-objectives,eq:attenuation-mae}. However, they are separate for different types of Gaussians.  We therefore combine image reconstruction with a constraint on the spectral energy distribution (SED), which
describes how flux varies with
wavelength~\cite{conroy2013panchromatic}.
For each view, we collect the GT log fluxes in
$\mathbf f=(f_1,\ldots,f_C)^{\mathsf T}$, where
$f_c=\log_{10}[\operatorname{sum}_c(\mathbf I^{\mathrm{GT}})]$
and $\operatorname{sum}_c$ sums all pixel intensities in band $c$.
The predicted vector $\widehat{\mathbf f}$ is constructed
analogously.
The SED loss is
\begin{equation}
\mathcal L_{\mathrm{SED}}
=\operatorname{MSE}\!\left(
\widehat{\mathbf f}-\widehat{\mu}_f,\,
\mathbf f-\mu_f\right).
\label{eq:sed-loss}
\end{equation}
where $\widehat{\mu}_f$ and $\mu_f$ are the means of
$\widehat{\mathbf f}$ and $\mathbf f$, respectively.
Finally, combining NMSE with the SED term gives the joint objective for refinement
\begin{equation}
  \mathcal L_{\mathrm{joint}}
    =\lambda_p\operatorname{NMSE}(\mathbf I,\mathbf I^{\mathrm{GT}})
      +\lambda_s\mathcal L_{\mathrm{SED}}.
\label{eq:joint-block-objectives}
\end{equation}

\section{Experiments}
\label{sec:experiments}

\noindent{\bf Datasets.} We use SKIRT-Auriga~\cite{kapoor2021auriga} and TNG50~\cite{costantin2023ceers} for evaluation. Auriga uses zoom simulations to study individual galaxies similar in mass to the Milky Way~\cite{grand2017auriga}. We select ten galaxies, each with 17 views. We use 14 views for training and 3 for testing. Each view provides a 50-band image, spanning $0.15$-$1244\,\mu\mathrm m$ with 21 bands in the UV-NIR group, 9 in the MIR group, and 20 in the FIR/submm group. The FIR/submm group includes the $22$-$24\,\mu\mathrm m$ band range. Training and evaluation use $512\times512$ images. TNG50 simulates a population of galaxies within a large region of the universe. We select ten galaxies, each with 14 views. We use 11 views for training and 3 for testing. Each view provides a 27-band image spanning $0.7$--$4.8\,\mu\mathrm m$, with all bands assigned to the UV-NIR group. Training and evaluation use $256\times256$ images. Note that in both datasets, the number of views is small, \ie a typical sparse view setting.

{\bf Baselines.} Since there are no directly comparable baselines, we compare \methodname{} with six methods with adaptations for galaxy data. These baselines include TensoRF-VM-MB~\cite{chen2022tensorf}, FeatSplat-MB~\cite{martins2024featsplat}, MS-Splatting~\cite{meyer2026mssplatting}, LGDWT-GS-MB~\cite{salehi2026lgdwt}, and 3DSatellites-MB~\cite{chang2025satellites}. In addition, we also augment the original 3DGS~\cite{kerbl2023gaussian} to 3DGS-MB, where we use degree-three spherical harmonic (SH) coefficients per band.

\begin{table*}[tb]
    \vspace{-1em}
    \caption{Comparison over three seeds on Auriga (top) and TNG50 (bottom). Red, orange, and yellow mark the best, second, and third.}
  \centering
  \small\rmfamily
  \setlength{\tabcolsep}{3pt}
  \begin{tabular*}{\textwidth}{@{\extracolsep{\fill}}l*{7}{c}@{}}
    \toprule
    Method & NMSE $\downarrow$ & $\mathrm{SAM}_w\downarrow$ & $E_{\mathrm{flux}}\downarrow$ & $E_{\mathrm{SED}}\downarrow$ & $E_{R_e}\downarrow$ & $\mathrm{SW}_1\downarrow$ & $E_{M_{\mathrm d}}\downarrow$ \\
    \midrule
    FeatSplat-MB~\cite{martins2024featsplat} & 0.516\textpm{}0.154 & 0.328\textpm{}0.041 & 0.336\textpm{}0.126 & 0.186\textpm{}0.089 & 0.121\textpm{}0.051 & 6.817\textpm{}3.613 & 0.107\textpm{}0.049 \\
    3DGS-MB~\cite{kerbl2023gaussian} & \yellowc 0.354\textpm{}0.114 & \yellowc 0.261\textpm{}0.057 & \yellowc 0.256\textpm{}0.072 & 0.145\textpm{}0.044 & \yellowc 0.094\textpm{}0.044 & \yellowc 4.789\textpm{}2.396 & \orangec 0.093\textpm{}0.035 \\
    LGDWT-GS-MB~\cite{salehi2026lgdwt} & 0.398\textpm{}0.146 & \orangec 0.258\textpm{}0.058 & \orangec 0.255\textpm{}0.084 & \yellowc 0.142\textpm{}0.042 & 0.095\textpm{}0.048 & \orangec 4.686\textpm{}2.340 & \yellowc 0.095\textpm{}0.044 \\
    MS-Splatting~\cite{meyer2026mssplatting} & 2.684\textpm{}5.069 & 0.630\textpm{}0.126 & 0.397\textpm{}0.106 & 0.342\textpm{}0.141 & 0.300\textpm{}0.144 & 12.216\textpm{}2.153 & 0.181\textpm{}0.118 \\
    TensoRF-VM-MB~\cite{chen2022tensorf} & 0.838\textpm{}0.492 & 0.478\textpm{}0.100 & 0.367\textpm{}0.195 & \orangec 0.140\textpm{}0.059 & 0.202\textpm{}0.144 & 10.218\textpm{}4.714 & 0.163\textpm{}0.097 \\
    3DSatellites-MB~\cite{chang2025satellites} & \orangec 0.158\textpm{}0.059 & 0.451\textpm{}0.079 & 0.321\textpm{}0.124 & 0.162\textpm{}0.049 & \orangec 0.082\textpm{}0.030 & 8.151\textpm{}3.444 & 0.172\textpm{}0.079 \\
    \midrule
    \methodname{} (ours) & \redc0.109\textpm{}0.029 & \redc0.190\textpm{}0.035 & \redc0.060\textpm{}0.021 & \redc0.068\textpm{}0.036 & \redc0.021\textpm{}0.008 & \redc1.429\textpm{}0.549 & \redc0.018\textpm{}0.005 \\
    \bottomrule
  \end{tabular*}
  \begin{tabular*}{\textwidth}{@{\extracolsep{\fill}}l*{6}{c}@{}}
    \toprule
    Method & NMSE $\downarrow$ & $\mathrm{SAM}_w\downarrow$ & $E_{\mathrm{flux}}\downarrow$ & $E_{\mathrm{SED}}\downarrow$ & $E_{R_e}\downarrow$ & $\mathrm{SW}_1\downarrow$ \\
    \midrule
    FeatSplat-MB~\cite{martins2024featsplat} & 1.033\textpm{}0.172 & \yellowc 0.411\textpm{}0.113 & 0.879\textpm{}0.225 & 0.193\textpm{}0.089 & 0.769\textpm{}0.199 & 42.565\textpm{}5.109 \\
    3DGS-MB~\cite{kerbl2023gaussian} & \yellowc 0.694\textpm{}0.180 & 0.419\textpm{}0.155 & \yellowc 0.705\textpm{}0.360 & \yellowc 0.175\textpm{}0.171 & \yellowc 0.505\textpm{}0.255 & \yellowc 21.566\textpm{}8.399 \\
    LGDWT-GS-MB~\cite{salehi2026lgdwt} & 0.710\textpm{}0.160 & 0.420\textpm{}0.159 & 0.747\textpm{}0.393 & 0.202\textpm{}0.205 & 0.518\textpm{}0.243 & 22.414\textpm{}8.255 \\
    MS-Splatting~\cite{meyer2026mssplatting} & 1.647\textpm{}0.865 & 0.546\textpm{}0.279 & 0.710\textpm{}0.489 & 0.212\textpm{}0.111 & 0.787\textpm{}0.197 & 45.417\textpm{}6.503 \\
    TensoRF-VM-MB~\cite{chen2022tensorf} & 1.480\textpm{}1.315 & \orangec 0.371\textpm{}0.205 & 0.996\textpm{}1.611 & 0.211\textpm{}0.297 & 0.622\textpm{}0.252 & 30.319\textpm{}8.264 \\
    3DSatellites-MB~\cite{chang2025satellites} & \orangec 0.372\textpm{}0.711 & 0.503\textpm{}0.285 & \orangec 0.313\textpm{}0.207 & \orangec 0.150\textpm{}0.103 & \orangec 0.153\textpm{}0.145 & \orangec 7.786\textpm{}3.757 \\
    \midrule
    \methodname{} (ours) & \redc0.313\textpm{}0.264 & \redc0.277\textpm{}0.095 & \redc0.129\textpm{}0.076 & \redc0.044\textpm{}0.027 & \redc0.078\textpm{}0.055 & \redc3.717\textpm{}0.947 \\
    \bottomrule
  \end{tabular*}
    \label{tab:comparison}
    \vspace{-1em}
\end{table*}
{\bf Metrics.} We assess two aspects for evaluation, \ie, reconstruction quality and physical fidelity. For all metrics, lower values indicate better results. 

To evaluate reconstruction quality, we use NMSE to measure reconstruction errors in pixel intensity relative to each band's GT scale. At each pixel, we also compare spectral similarity using the spectral angle mapper (SAM)~\cite{kruse1993sips}, $\mathrm{SAM}_w$.
For physical fidelity, we use the flux error $E_{\mathrm{flux}}$ to compare each band's total flux with GT. Additionally, the SED error $E_{\mathrm{SED}}$ compares the relative band flux distribution. To validate the spatial difference in astronomical images, we employ the Sliced Wasserstein-1 distance ($\mathrm{SW}_1$) to measure the predicted and GT light distributions~\cite{kolouri2019generalized}. To measure how far the light extends, we compare the difference between the predicted and GT images' half-light radii, the semimajor axes of ellipses enclosing half of each image's total light~\cite{baes2024radius}, denoted as $E_{R_e}$. 
We also compare dust mass differences $E_{M_{\mathrm d}}$ in the FIR range, estimated from the predicted and GT images. We evaluate this on Auriga, as TNG50 lacks data in the FIR range.

\subsection{Experimental Results}
\label{sec:quantitative-comparison}
\subsubsection{Quantitative comparison}

\Cref{tab:comparison} compares \methodname{} with six baselines on Auriga and TNG50, where \methodname{} achieves the lowest mean error on all metrics. While on visual similarity metrics such as NMSE, GalaxyGS already universally improves the results, its improvements in physical fidelity metrics are even more prominent, often by large margins, which is particularly useful in demonstrating that GalaxyGS is not only seeking visual similarity in reconstruction. 

On Auriga, \methodname{} achieves an $\mathrm{SW}_1$ of $1.429$, compared with $4.686$ for the next-best LGDWT-GS-MB, which employs wavelet losses to preserve spatial structure and has an error $227.9\%$ higher than ours. Our separate stellar and dust emission fields allow wavelength-dependent spatial distributions, compared with LGDWT-GS-MB's single field. In addition, we also achieve an $E_{\mathrm{SED}}$ of $0.068$, compared with $0.140$ for the next-best TensoRF-VM-MB, whose error is $106.9\%$ higher than ours. While TensoRF-VM-MB predicts all bands from a factorized field, our reference PCA bases constrain cross-band intensity variations within each emission field.

On TNG50, our method achieves an $E_{\mathrm{SED}}$ of $0.044$, compared with $0.150$ from the next-best method 3DSatellites-MB, whose error is $244.5\%$ higher than ours. While 3DSatellites-MB extends its Gaussian representation to multiple bands, our model explicitly separates dust attenuation from stellar emission. This enables wavelength-dependent attenuation along each viewing ray, accounting for both intensity reduction and changes in spectral shape. 

\subsubsection{Qualitative comparison}
\begin{figure*}
   \vspace{-1em}
  \centering
  \includegraphics[width=\textwidth]{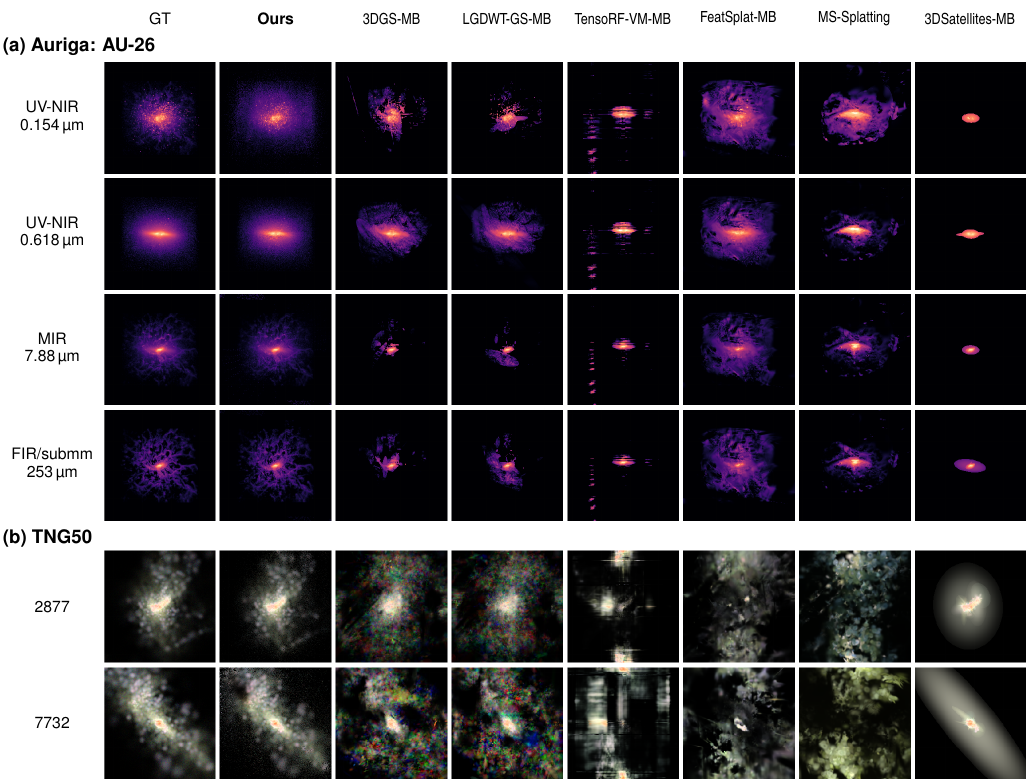}
  \caption{Light-intensity comparison for Auriga AU-26 (top) and three-color comparisons for TNG50 galaxies 2877 and 7732 (bottom), each at a test view. For TNG50, the $4.4$, $3.57$, and $2.78\,\mu\mathrm{m}$ bands are mapped to red, green, and blue, respectively.}
  \label{fig:au-band-comparison}
  \label{fig:matched-qualitative}
  \vspace{-1em}
\end{figure*}
\label{sec:qualitative-comparison}
The aggregate metrics above cannot directly demonstrate the scale or prominence of the difference. Therefore, we also show qualitative results. For Auriga, we show four bands from one test view of galaxy AU-26 in \cref{fig:au-band-comparison}. In AU-26, TensoRF-VM-MB and 3DSatellites-MB recover the bright disk but completely miss the infrared light above it. In contrast, FeatSplat-MB recovers dust emission, but underestimates the bright disk. All other baseline methods are in between. \methodname{} retains the bright disk together with much of the surrounding emission. The outer infrared emission of AU-26 provides a second example. LGDWT-GS-MB recovers the bright center, with disconnected patches in the surrounding infrared light. 3DSatellites-MB produces a compact center with little of the extended structure. \methodname{} retains the compact center and the wider infrared details. 

The three-color TNG50 views show both bright structures and surrounding diffuse emission. 3DGS-MB and LGDWT-GS-MB are best among the baselines but still produce colored artifacts around bright regions, while 3DSatellites-MB oversmooths the diffuse emission, reducing the visibility of individual bright structures. In contrast, \methodname{} preserves the curved central structure in galaxy 2877 and the diagonal distribution of bright regions in galaxy 7732, while retaining the surrounding diffuse emission.

\begin{figure}[t]
  \vspace{-1em}
  \centering
  \includegraphics[width=\linewidth]{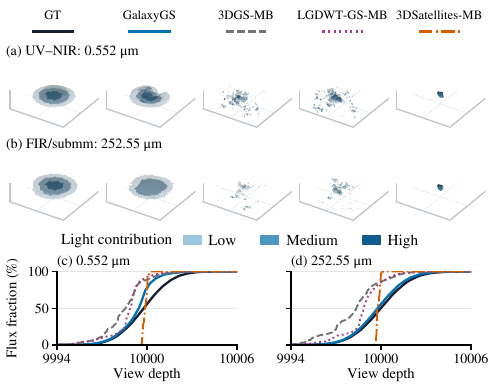}
  \caption{Top: 3D light distributions in two bands, with darker surfaces indicating greater light contributions. Bottom: cumulative fraction of image flux from positions closer to the observer. View depth increases away from the observer.}
  \label{fig:controlled-light-distribution}
\end{figure}

We further evaluate physical fidelity using a controlled scene. We simulated a galaxy where we have ground truth for emission and attenuation separately, so that predictions can be compared with the true emission of astronomical bodies.  \Cref{fig:controlled-light-distribution} compares the 3D light distributions and cumulative flux along the viewing direction. In the top panels of~\cref{fig:controlled-light-distribution}, \methodname{} produces distributions more similar to GT. In contrast, 3DGS-MB and LGDWT-GS-MB produce fragmented distributions shifted toward the observer, while 3DSatellites-MB concentrates flux within a narrow depth range. In the bottom panels of~\cref{fig:controlled-light-distribution}, the curves show the percentage of total band flux contributed by objects closer to the observer across depths. Our cumulative flux profiles better match the gradual increase observed in GT, indicating a more physically accurate reconstruction of the spatial distribution of light.

\subsection{Ablation Studies}
\label{sec:ablation-studies}
We test several key model designs and their alternatives. To show whether stellar and dust emission should be separately modeled, we make them share means $\boldsymbol{\mu}$ and covariances $\boldsymbol{\Sigma}$ while keeping their emission learning separate, in `Shared $\boldsymbol\mu,\boldsymbol\Sigma$'. Next, to test the necessity of using PCA bases to constrain the learning of stellar and dust emission, we replace the emission intensity $e_c$ in \eqref{eq:emission-decoder} with an independently learned nonnegative scalar for each Gaussian and band, in `Direct stellar $e_c$' and `Direct dust $e_c$', respectively.  Also, to test the benefit of sharing means and covariances among MIR and FIR/submm Gaussians, we separate them in `Separate MIR/FIR $\boldsymbol\mu,\boldsymbol\Sigma$'.

Also, to test the modeling choice for attenuation, we present three alternatives: removing the attenuation field in `No attenuation field', a single $\tau$ for all UV-NIR bands per Gaussian in `One $\tau$ across bands', and replacing the RBF parameterization with an independently learnable $\tau_c$ per band in `Independent $\tau_c$'. In addition, to show the benefits of staged optimization, we test training all types of Gaussians simultaneously in `Joint optimization'. 

\begin{table}
  \vspace{-1em}
  \caption{Ablation study on AU-26. Bold indicates the lowest error.}
  \label{tab:ablation-components}
  \centering
  \small\rmfamily
  \setlength{\tabcolsep}{1.25pt}
  \begin{tabular*}{\columnwidth}{@{\extracolsep{\fill}}lcccc@{}}
    \toprule
    Variant & NMSE $\downarrow$ & $E_{\mathrm{flux}}\downarrow$ & $E_{\mathrm{SED}}\downarrow$ & $\mathrm{SW}_1\downarrow$ \\
    \midrule
    Shared $\boldsymbol\mu,\boldsymbol\Sigma$ & 0.203 & 0.265 & 0.140 & 2.451 \\
    Direct stellar $e_c$ & 0.167 & 0.166 & 0.197 & 2.217 \\
    Direct dust $e_c$ & 0.235 & 0.141 & 0.115 & 1.332 \\
    Separate MIR/FIR $\boldsymbol\mu,\boldsymbol\Sigma$ & 
    \textbf{0.060} & 0.035 & 0.034 & \textbf{0.494} 
    \\
    \addlinespace[2pt]
    No attenuation field & 0.069 & 0.036 & 0.035 & 0.566 \\
    One $\tau$ across bands & 0.069 & 0.035 & 0.035 & 0.564 \\
    Independent $\tau_c$ & 0.068 & 0.034 & 0.038 & 0.536 \\
    \addlinespace[2pt]
    Joint optimization & 0.152 & 0.113 & 0.130 & 1.758 \\
    \midrule
    Full model & 0.067 & \textbf{0.032} & \textbf{0.032} & 0.530 \\
    \bottomrule
  \end{tabular*}
  \vspace{-1em}
\end{table}

We show the overall results of all variants of our model in~\cref{tab:ablation-components}. First, `Shared $\boldsymbol\mu,\boldsymbol\Sigma$' gives the worst overall results, showing the need for separate modeling for stellar and dust emission. This is not surprising as they come from distinctive astronomical bodies, differing in their emission mechanisms and spatial distributions. Next, unconstrained learning for stellar and dust emission leads to much worse results, as shown by `Direct stellar $e_c$' and `Direct dust $e_c$', especially for metrics related to physical fidelity. Direct fitting does not constrain the learning within the physical PCA basis. Theoretically, it can learn the same emission fields as GalaxyGS, but fails to do so in practice. We speculate that this is because independently learning induces additional local minima, so that the learning does not need to stay in a physically meaningful space. Finally, `Separate MIR/FIR $\boldsymbol\mu,\boldsymbol\Sigma$' can achieve better NMSE and $\mathrm{SW}_1$, but it is worse on $E_{\mathrm{flux}}$ and $E_{\mathrm{SED}}$, so we keep it as a close alternative.

Next, `No attenuation field' shows that emission fields, if modeled well, can achieve relatively good results. However, if the attenuation field is not modeled well, such as `One $\tau$ across bands' and `Independent $\tau_c$', it cannot further improve the results.  GalaxyGS further improves the results.

Finally, even if all three Gaussian fields are modeled well, jointly optimizing them will suffer from local minima, as in `Joint optimization'. This is mainly due to the uncontrolled compensation between emissions and attenuation.

\section{Conclusion, Discussion, and Future Work}
\label{sec:conclusion}

We presented \methodname{}, the first 3DGS method for multi-band galaxy novel view synthesis. By turning 3DGS into a physically inspired formulation, \methodname{} represents stellar emission, dust attenuation, and dust emission, models light propagation and attenuation, and trains via staged optimization. Experiments demonstrate that \methodname{} outperforms all baselines, often by large margins. 

However, while the physical priors constrain the spectra of individual fields, \methodname{} does not explicitly link the starlight absorbed by dust to its infrared emission. Also, reconstruction requires multiple views of the same galaxy to constrain its 3D structure, limiting its application to galaxies observed from a single direction. Future work will improve the physical fidelity by coupling dust absorption and emission and explore structural priors that could enable single-view reconstruction.

\section*{Acknowledgement}
This work was supported in part by the UK Research and Innovation AIRR Gateway Award
\mbox{(0261-5599-0335-1)}.

{\small
  \bibliographystyle{ieeenat_fullname}
  \bibliography{main}
}
\end{document}